\documentclass[lettersize,journal]{IEEEtran}
\usepackage{amsmath,amsfonts}
\usepackage{algorithm}
\usepackage{algorithmic}
\usepackage{array}
\usepackage[caption=false,font=normalsize,labelfont=sf,textfont=sf]{subfig}
\usepackage{textcomp}
\usepackage{stfloats}
\usepackage{url}
\usepackage{verbatim}
\usepackage{url}            
\usepackage{booktabs}       
\usepackage{amsfonts}       
\usepackage{nicefrac}       
\usepackage{microtype}      
\usepackage{xcolor}         
\usepackage{amsthm}
\usepackage{amsopn}
\usepackage{amsmath}
\usepackage{graphicx}
\usepackage{booktabs}
\usepackage{calc}
\usepackage{amsthm}
\usepackage{siunitx}
\usepackage{hyperref}       
\hypersetup{
    pdfborder={0 0 0}
}

\newtheorem{lemma}{Lemma}

\providecommand{\R}{{\mathbb R}}

\providecommand{\E}{{\mathbb E}}

\providecommand{\BA}{{\mathbf A}}

\providecommand{\BZ}{{\mathbf Z}}

\providecommand{\Bx}{{\mathbf x}}

\providecommand{\By}{{\mathbf y}}

\DeclareMathOperator*{\argmin}{arg\,min}

\usepackage{enumitem,amssymb}
\newlist{todolist}{itemize}{2}
\setlist[todolist]{label=$\square$}

\def\BibTeX{{\rm B\kern-.05em{\sc i\kern-.025em b}\kern-.08em
    T\kern-.1667em\lower.7ex\hbox{E}\kern-.125emX}}
\usepackage{balance}
\begin{document}
\title{Learning Binary Sampling Patterns for Single-Pixel Imaging using Bilevel Optimisation}
\author{Serban Cristian Tudosie$^*$, Alexander Denker, \v{Z}eljko Kereta, Simon Arridge
\thanks{S.C.T. acknowledges financial support from the UK Research and Innovation (UKRI) under the UK government’s Horizon Europe funding Guarantee EP/X030733/1 and the European Union GA101072354. A.D. acknowledges support from the EPSRC programme grant EP/V026259/1. Z.K. acknowledges support by the UK EPSRC grant EP/X010740/1. S.C.T., A.D., Z.K., and S.A are with the Department of Computer Science, University College London. \\ $^*$corresponding author sc.tudosie@ucl.ac.uk}}

\maketitle
\begin{abstract}
Single-Pixel Imaging (SPI) enables the reconstruction of objects using a single detector through sequential illuminations with structured light patterns. The choice of illumination patterns is critical, particularly in highly undersampled regimes, where it directly determines reconstruction quality and acquisition speed. Instead of relying on handcrafted or fixed patterns, we propose to learn task-specific patterns directly from data. Practical SPI hardware only supports binary patterns, making binary pattern design a necessary consideration. We propose a bilevel optimisation method for learning task-specific binary illumination patterns optimised for applications such as single-pixel fluorescence microscopy. We address the non-differentiable nature of binary optimisation using the Straight-Through Estimator. In addition, we incorporate learned variational regularisation, improving reconstruction quality and robustness. We demonstrate our method on the CytoImageNet microscopy dataset. We show that our learned patterns achieve superior reconstruction performance compared to baseline methods and end-to-end deep learning, particularly in highly undersampled regimes and in scarce-data settings. 
\end{abstract}


\section{Introduction}
\begin{figure}[t]
    \centering
    \includegraphics[width=0.95\linewidth]{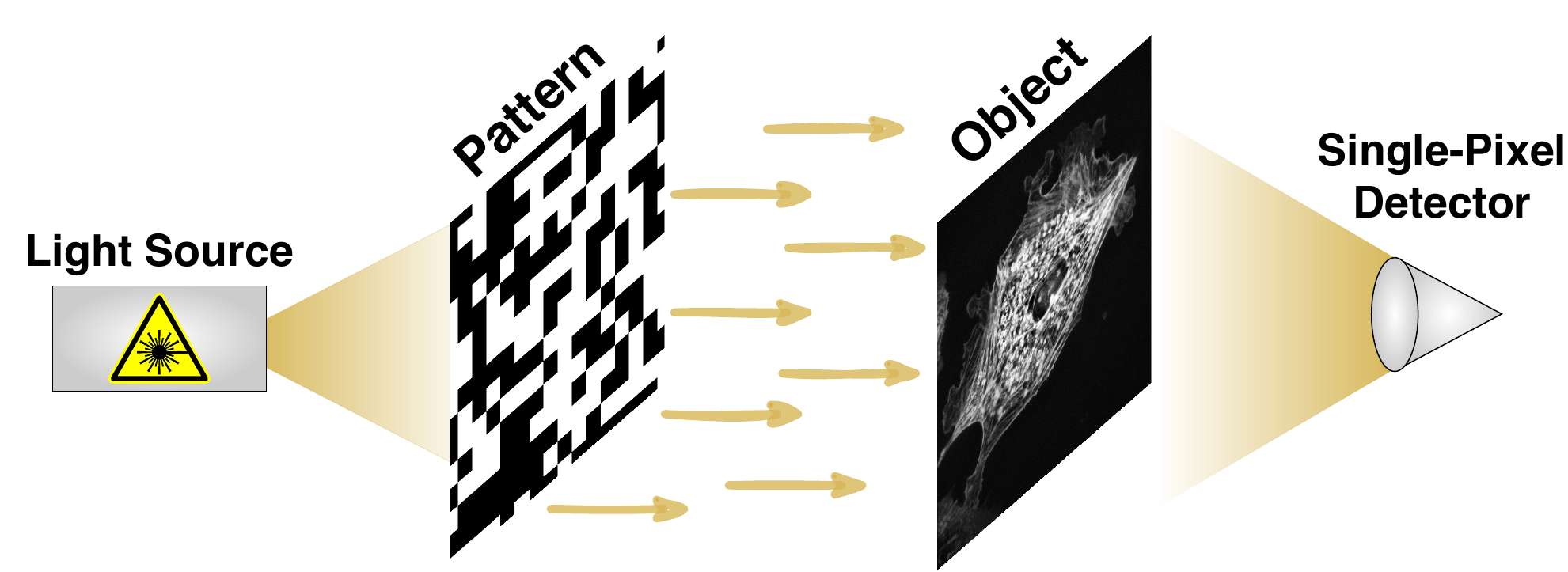}
    \caption{Simplified schematic of the forward process in SPI. For clarity, only one illumination pattern is illustrated.}
    \label{fig:spi}
\end{figure}

\IEEEPARstart{S}{ingle-Pixel} Imaging (SPI) is an imaging technique that uses a single photodetector to measure the total intensity of light transmitted under a given illumination pattern \cite{Duarte_2008_SinglepixelImagingCompressive, Edgar_2019_PrinciplesProspectsSinglepixel}. 
Since each measurement outputs a single scalar value, SPI is inherently non-spatially resolving and, therefore, cannot directly determine where the detected light originated.
Spatial information is instead resolved by illuminating the object with a sequence of structured light patterns, see Fig.~\ref{fig:spi}. 

Relying on a single bucket detector rather than a detector array is cost-effective and particularly well suited to biomedical applications, as it lowers the cost of multispectral, hyperspectral, and fluorescence-lifetime imaging~\cite{Ghezzi_2023_ComputationalBasedTimeresolved, Uguen_2024_SinglepixelbasedHyperspectralMicroscopy, Li_2017_EfficientSinglepixelMultispectral}.
Furthermore, SPI can provide better signal-to-noise ratio performance under photon-limited conditions compared to array sensors, and single-pixel detectors can operate in spectral regions, such as infrared or terahertz bands, where conventional imaging systems require costly cameras~\cite{Li_2017_EfficientSinglepixelMultispectral}.

Data acquisition can be described by the linear forward model 
\begin{align}
    \label{eq:forward_op}
    \By^\delta = P( \By), \quad y_i = \langle \BA_i, \Bx \rangle,
\end{align}
where $\Bx \in \R^N$ represents the (vectorised) object, $\BA \in \{-1, 1\}^{M \times N}$ is the sampling matrix, with rows representing illumination (sampling) patterns, $\By^\delta \in \R^M$ are the (noisy) measurements, and $P$ is some noising process.
The object is then reconstructed using the sequence of recorded measurements $\By^\delta$ along with the known illumination patterns~$\BA$.

A central challenge in SPI is determining which illumination patterns to use and how many are required to achieve a reconstruction of adequate quality. Since the acquisition time scales directly with the number of projected patterns $M$, reducing $M$ is essential for enabling high-speed imaging.

The choice and design of illumination patterns are constrained by practical SPI hardware.
Spatial modulation of light is most commonly achieved using digital micromirror devices, which consist of microscopic mirrors that direct light either towards a lens (``on'') or towards a light absorber (``off''), producing bright and dark pixels. 
As a result, practical SPI implementations are limited to sampling patterns with unipolar binary values $\{0,1\}$. 
To obtain a better-conditioned measurement matrix, unipolar patterns are often coupled via sequential subtraction to emulate bipolar binary patterns $\{-1,1\}$~\cite{Wu_2022_SequentialSubtractionBasedCompressive}.

Hadamard patterns are one of the most well-studied classes of binary $\{-1,1\}$ sampling patterns in SPI~\cite{Gibson_2020_SinglepixelImaging12a, Yu_2020_SuperSubNyquistSinglePixel}. 
They form an orthogonal basis and exhibit a hierarchical, frequency-like ordering that enables progressive image reconstruction, where coarse object structures emerge early in the acquisition process. 
However, in low-sampling regimes (high compression, low $M$), they often create severe blocky artefacts in reconstructed images.
Scrambled Hadamard (SH) patterns address this limitation by randomly permuting the rows and columns of the Hadamard matrix~\cite{Huynh_2019_SinglepixelCameraPhotoacoustic}. 
This preserves orthogonality while increasing incoherence with sparsifying bases, thereby improving compressed sensing performance. 
The reconstruction performance of SPI systems depends not only on the choice of the pattern family but also on the selection strategy used to rank and apply a limited subset of patterns.
For Hadamard-based SPI, several ordering (ranking) strategies have been proposed, including TV-ordering~\cite{Yu_2020_SuperSubNyquistSinglePixel}, CG-ordering~\cite{Yu_2019_SuperSubNyquistSinglePixel}, and XY-ordering~\cite{Cai_2023_DetailenhancedSamplingStrategy}, each yielding different behaviour in low-sampling regimes. Furthermore, sequential sampling methods have been developed for Fourier SPI; see, e.g., \cite{Yuan_2021_AdaptiveDynamicOrdering}. However, such strategies exploit the structure of the Fourier space, and are not directly extendable to binary $\{-1,1\}$ patterns. 

In applications such as single-pixel fluorescence microscopy, a key challenge is the low spatial resolution imposed by optical limitations~\cite{Ebner_2023_DiffractionlimitedHyperspectralMidinfrared}. 
In such systems, the single-pixel detector acquires low-resolution measurements across multiple spectral bands and fluorescence decay time bins.
These measurements are then fused with a high-resolution single-channel intensity image using data fusion techniques~\cite{Jerez_2018_SinglePixelSpectral, Soldevila_2021_GigavoxelMultidimensionalFluorescencea, Ghezzi_2023_ComputationalBasedTimeresolved, Simoni_2026_Computational3DMultispectral}, to recover a high-resolution tensor. 
However, reliance on Hadamard or SH patterns limits the acquisition speed, motivating the need for alternative sampling pattern designs.

We propose a data-driven approach that \emph{learns} illumination patterns to optimise the reconstruction performance on a given dataset by adapting to the statistics of the target objects. 
In particular, given a dataset $\mathcal{D}= \{ \Bx^{(i)} \}_{i=1}^n$ of images and a reconstruction operator $\mathcal{R}: \R^M \to \R^N$, we learn the sampling patterns, represented as $\BA \in \{-1,1\}^{M \times N}$, by minimising the empirical reconstruction error
\begin{align}
    \label{eq:pattern_learning}
    \min_{\BA \in \{-1,1\}^{M \times N}} \frac{1}{n} \sum_{i=1}^n \mathcal{L}(\Bx^{(i)}, \mathcal{R}(P(\BA \Bx^{(i)}))),
\end{align}
where $\mathcal{L}(\Bx, \hat{\Bx}) = \| \Bx - \hat{\Bx}\|^2$ denotes the mean squared error. 
Learning binary sampling patterns for SPI, instead of relying on fixed designs, was introduced by DCAN~\cite{Higham_2018_DeepLearningRealtime}, and was later extended to ternary $\{-1,0,1\}$ patterns by DLSVD~\cite{Deng_2024_SinglePixelImagingBased}. 

However, three fundamental questions remain:
\begin{enumerate}[noitemsep]
    \item how to optimise over the discrete space of binary matrices;
    \item which reconstruction operator $\mathcal{R}$ to use;
    \item how to relate the pattern optimisation to reconstruction performance.
\end{enumerate}
The binary nature of the patterns renders the optimisation both non-convex and non-differentiable, precluding the direct application of gradient-based methods. 
In particular, the search space is combinatorial, of size $2^{MN}$, making exhaustive exploration intractable even for moderate image resolution. 
A common approach to address this is to relax the binary constraint to $\BA \in [-a,a]^{M \times N}, a\ge1$, and to introduce a penalty term in \eqref{eq:pattern_learning} to drive the patterns towards $\{-1,1\}$ during training. 
A common example of this approach is \textit{Relax-and-Penalise} (RnP)~\cite{Lucidi_2010_ExactPenaltyFunctions} where the penalty function is
\begin{align}
    \label{eq:rnp}
    r_\epsilon(\BA) = \frac{1}{\epsilon}\sum_{i,j} (1-A_{ij})(1+A_{ij}),\ \  \BA \in [-1,1]^{M \times N},
\end{align} 
with $a = 1$.
Similarly, DCAN, and its extensions, set $a = \infty$ and the penalty function to a multi-well potential
\begin{align}
\label{eq:multi_well}
    r_\epsilon(\BA) = \frac{1}{\epsilon} \sum_{i,j} (1-A_{ij})^2(1+A_{ij})^2, \ \  \BA \in \R^{M \times N}.
\end{align}
However, these approaches require careful tuning of the penalty schedule $\epsilon$ and do not guarantee strict binarisation during training~\cite{Lucidi_2010_ExactPenaltyFunctions}.
Furthermore, DCAN and its extensions jointly optimise the sampling patterns alongside a neural network reconstructor $\mathcal{R}_\theta$, leading to the joint optimisation problem
\begin{align}
    \label{eq:joint_pattern_learning}
    \min_{\substack{\BA \in \{-1,1\}^{M \times N} \\ \theta \in \Theta}} \frac{1}{n} \sum_{i=1}^n \mathcal{L}(\Bx^{(i)}, \mathcal{R}_\theta(P(\BA \Bx^{(i)}))).
\end{align}
However, this joint approach demands large training sets, i.e., on the order of \num{29000} to \num{45000} images~\cite{Wang_2022_SinglepixelImagingUsing, Deng_2024_SinglePixelImagingBased}.

\textbf{Contributions.} We propose two key modifications of this data-driven paradigm. First, we adopt the \textit{Straight-Through-Estimator} (STE) \cite{Bengio_2013_EstimatingPropagatingGradients} to optimise directly over binary patterns throughout training. Namely, patterns are represented by a latent real-valued matrix $\mathbf{Z} \in \mathbb{R}^{M \times N}$ as $\mathbf{A} = \texttt{sgn}(\mathbf{Z})$, and a differentiable surrogate is used to approximate the gradient during the backward pass. Second, rather than training a reconstructor jointly with the patterns, we employ a fixed, pre-trained reconstructor based on a learned regularisation functional, thereby decoupling pattern optimisation from network training. 
We employ two pre-trained learned regularisers: Total Deep Variation (TDV)~\cite{Kobler_2020_TotalDeepVariation} and the Weakly Convex Ridge Regularizer (WCRR)~\cite{Goujon_2023_LearningWeaklyConvex}, that provide powerful general-purpose priors that transfer well across datasets and measurement setups. 
Crucially, this decoupling dramatically reduces the data requirements for pattern learning.
We demonstrate competitive reconstruction performance using as few as $250$ training images, compared to tens of thousands required by previous approaches. We further investigate the impact of binarisation strategy, STE surrogate choice, and measurement noise, providing a comprehensive framework for robust data-driven pattern design in SPI.


The paper is structured as follows. 
Section~\ref{sec:method} details our bilevel training framework for learning binary patterns. 
Section~\ref{sec:prob_inter_ste} provides a probabilistic interpretation of the STE. 
In Section~\ref{sec:numerical_exp} we present numerical results on the CytoImageNet dataset \cite{Hua_2021_CytoImageNetLargescalePretraining}.
In Section~\ref{sec:conc} we provide conclusions and an outlook for further work.

\subsection{Related Work}

\paragraph{Data-driven Pattern Design} Joint optimisation of illumination patterns and the reconstruction network for SPI was introduced by DCAN~\cite{Higham_2018_DeepLearningRealtime}.
There, the imaging process is formulated as an autoencoder: the encoder is the linear measurement model \eqref{eq:forward_op} and the decoder is a deep neural network trained end-to-end for image reconstruction. 
Wang et al.~\cite{Wang_2022_SinglepixelImagingUsing} build on this approach with a physics-inspired decoder that first applies differential ghost imaging~\cite{Ferri_2010_DifferentialGhostImaging} as a feature-extraction layer before refining the result with a learned network.
DLSVD~\cite{Deng_2024_SinglePixelImagingBased} extends the framework to ternary $\{-1,0,1\}$ patterns and employs an SVD-based linear reconstruction stage.

\paragraph{Magnetic Resonance Imaging (MRI)}
Closely related work appears in accelerated MRI, where the goal is to select Fourier coefficients under a fixed sampling budget.
Bilevel learning has been widely explored in this context~\cite{Sherry_2020_LearningSamplingPattern,Baldassarre_2016_LearningBasedCompressiveSubsampling,Gozcu_2018_LearningBasedCompressiveMRI,Ravula_2023_OptimizingSamplingPatterns}, including extensions to \emph{sequential} (adaptive) sampling, where each measurement is chosen based on an intermediate reconstruction~\cite{Yin_2021_EndtoEndSequentialSampling}.

\paragraph{Binary Neural Networks} STE is a widely used technique for training neural networks with discrete or binary (quantised) weights, where gradients are otherwise undefined. Originally introduced for stochastic and binary neurons by Bengio et al.~\cite{Bengio_2013_EstimatingPropagatingGradients},  STE replaces the non-differentiable quantisation operation in the backward pass with a surrogate gradient. This enables end-to-end optimisation with standard gradient-based methods for discrete or binary neural networks, see e.g. \cite{courbariaux2015binaryconnect,hubara2016binarized}. In this work, we adopt the STE to enable gradient-based optimisation of binary $\{-1,1\}$ measurement patterns.

\section{Method}
\label{sec:method}

\subsection{Problem Formulation}
We seek to learn an optimal set of binary illumination patterns, organised as  $\BA \in \{-1,1\}^{M \times N}$, such that the resulting reconstructions minimise the reconstruction error over a representative dataset of images $\{ \Bx^{(i)} \}_{i=1}^n$.
Here, we define the reconstructor using the variational regularisation framework with a pre-trained regularisation functional $\mathcal{J}$.
Under this choice, the optimisation problem in \eqref{eq:pattern_learning} is naturally cast as a \emph{bilevel optimisation} problem~\cite{Colson_2007_OverviewBilevelOptimization}. 
The problem formulation then reads
\begin{align}
\label{eq:bilevel_problem1}
    \min_{\substack{\BZ \in \mathbb{R}^{M \times N}\\ \alpha >0}} &\!\bigg\{ \mathcal{F}(\BZ,\!\alpha)\!=\!\sum_{i=1}^n \mathcal{L}\Big(\Bx^{(i)},\!\hat{\Bx}\big(\texttt{sgn}(\BZ), \alpha;\By^{(i)}\big)\Big) \bigg\},\\ 
\label{eq:bilevel_problem2}
    \text{s.t.}\ \ \hat{\Bx}&(\BA, \alpha;\By) \in \argmin_{\Bx\in \mathbb{R}^N} \frac{1}{2} \| \BA \Bx - \By \|_2^2 + \alpha  \mathcal{J}(\Bx),
\end{align}
where $\BA = \texttt{sgn}(\BZ)$, and $\By^{(i)} = P(\BA \Bx^{(i)})$ denotes simulated measurements under a noise process $P$. 
The upper-level problem \eqref{eq:bilevel_problem1} aims to find a latent representation $\BZ\in\mathbb{R}^{M\times N}$ and a regularisation weight $\alpha>0$. 
Reconstructions are obtained by solving the lower-level problem \eqref{eq:bilevel_problem2}, governed by the discrete matrix $\BA$ and $\alpha$.  
The matrix $\BZ \in \mathbb{R}^{M \times N}$ serves as a continuous (latent) relaxation of $\BA$, with $\BA = \texttt{sgn}(\BZ)$ applied element-wise. 
While the regularisation functional $\mathcal{J}$ is fixed, we tune its influence through the regularisation weight $\alpha$. 
 
The full optimisation procedure is summarised in Algorithm~\ref{alg:bilevel_opt}, and we now describe its key components.
To solve the bilevel problem \eqref{eq:bilevel_problem1}-\eqref{eq:bilevel_problem2} using gradient-based methods, we must compute the gradient $\nabla_\theta \mathcal{F}$, where $\theta = (\BZ, \alpha)$. Applying the chain rule for a single sample ($n=1$), we obtain
\begin{align}
    \nabla_\theta \mathcal{F}(\theta) = \Big[\frac{\partial \hat{\Bx}(\theta)}{\partial \theta}\Big]^T \nabla_{\Bx} \mathcal{L}(\hat{\Bx}(\theta)).
\end{align}
This main bottleneck lies in the the Jacobian $\frac{\partial \hat{\Bx}(\theta)}{\partial \theta}$ of the solution $\hat{\Bx}(\theta)$ to the lower-level problem \eqref{eq:bilevel_problem2}. We obtain the lower-level solution $\hat{\Bx}(\theta)$ via fixed-point iterations $\Bx^{(k+1)}=T_\theta(\Bx^{(k)})$ with a suitable operator $T_\theta$ defined by a reconstruction scheme.  Differentiating the fixed-point equation $\hat{\Bx}(\theta) = T_\theta(\hat{\Bx}(\theta))$ yields
\begin{align}
    \frac{\partial \hat{\Bx}(\theta)}{\partial \theta} = \left(\mathbf{I} - \frac{\partial T_\theta(\hat{\Bx}(\theta))}{\partial \hat{\Bx}(\theta)} \right)^{-1} \frac{\partial T_\theta(\hat{\Bx}(\theta))}{\partial \theta}.
    \label{eq:jacobian}
\end{align} 
There are several options for computing the Jacobian \eqref{eq:jacobian}. 
A classical strategy is to implicitly differentiate the optimality conditions of the lower-level problem with respect to $\theta$. 
However, this requires solving an additional linear system, whose conditioning depends on the current iterate and changes at every iteration of the bilevel formulation. In turn, this can lead to computationally expensive iterations \cite{Bai_2019_DeepEquilibriumModels}.
An alternative is to unroll the optimisation \cite{maclaurin2015gradient}. However, backpropagating through a long sequence of iterations and retaining the corresponding computational graph for large $k$ in $T_\theta({\Bx}^{(k)})$ quickly becomes prohibitive in memory. 
In practice, when learning the patterns $\BA$, the lower-level solver requires over a thousand iterations to converge, especially in the initial stages.

To reduce the computational cost, we use Jacobian-Free Backpropagation (JFB)~\cite{Fung_2022_JFBJacobianFreeBackpropagation}, which relies on a zero-order Neumann series approximation of the inverse term
\begin{align}
    \left( \mathbf{I} - \frac{\partial T_{\theta}(\mathbf{\hat{x}}(\theta))}{\partial \mathbf{\hat{x}}(\theta)} \right)^{-1} \approx \ \mathbf{I}.
    \label{eq:jfb}
\end{align}
This yields an approximate gradient that is computationally efficient and works well in practice \cite{Shaban_2019_TruncatedBackpropagationBilevel,Zou_2023_DeepEquilibriumLearning,Hertrich_2025_LearningRegularizationFunctionals}.

\begin{algorithm}[t]
\caption{Bilevel Optimisation of Binary Sampling Matrices}
\label{alg:bilevel_opt}
\begin{algorithmic}[1]
\REQUIRE Dataset $\{\Bx^{(i)}\}_{i=1}^{n}$, step sizes $\eta_{\BZ},\eta_{\alpha}$, batch size $B$
\REQUIRE Regulariser $\mathcal{J}$
\STATE Initialise $\BZ,\alpha$
\WHILE{not converged} 
\STATE Sample mini-batch $\{ \Bx^{(j)} \}_{j=1}^B$
\STATE Binary sampling matrix: $\BA = \texttt{sgn}(\BZ)$
\FOR{$j=1,\ldots,B$}
\STATE Measurements: $\By^{(j)} = P(\BA \Bx^{(j)})$
\STATE Reconstruction of variational problem:
\[
\hat{\Bx}^{(j)} \in 
\argmin_{\Bx} 
\frac{1}{2}\|\BA\Bx-\By^{(j)}\|_2^2 + \alpha \mathcal{J}(\Bx)
\]
\STATE Obtain Lipschitz estimate $L_j$
\ENDFOR
\STATE Upper level loss: $\mathcal{F}=\frac1B\sum_{j=1}^B\mathcal{L}(\Bx^{(j)},\hat{\Bx}^{(j)})$
\STATE Define one gradient step
{ \small
\[
T_{\theta}(\hat{\Bx}^{(j)}) =
\hat{\Bx}^{(j)} -
\frac{1}{L_j}
\Big(
\BA^T(\BA\hat{\Bx}^{(j)}-\By^{(j)}) + \alpha \nabla_{\Bx}\mathcal{J}(\hat{\Bx}^{(j)})
\Big)
\]}
\STATE Jacobian-free gradient approximation
\[
\nabla_{\theta}\mathcal{F}
\approx
\left[
\frac{\partial T_{\theta}(\hat{\Bx})}{\partial \theta}
\right]^T
\nabla_{\hat{\Bx}}\mathcal{L},
\quad
\theta=(\BA,\alpha)
\]
\STATE \textbf{STE:} $\nabla_{\BZ}^{\text{STE}}\mathcal{F}=g^{\text{STE}}(\BZ)\odot\nabla_{\BA}\mathcal{F}$
\STATE $\BZ\leftarrow\BZ-\eta_{\BZ}\nabla_{\BZ}^{\text{STE}}\mathcal{F}$,\quad
$\alpha\leftarrow\alpha-\eta_{\alpha}\nabla_{\alpha}\mathcal{F}$
\ENDWHILE
\RETURN $\BA^\star=\texttt{sgn}(\BZ),\ \alpha^\star$
\end{algorithmic}
\end{algorithm}

Once the gradient with respect to the discrete matrix $\nabla_\BA \mathcal{F}$ is computed, we must backpropagate it to the continuous latent matrix $\BZ$. Applying the chain rule yields
\begin{align}
    \nabla_\BZ \mathcal{F}(\BZ, \alpha) &= \texttt{sgn}'(\BZ) \odot \nabla_\BA \mathcal{F}(\BA, \alpha)
    \label{eq:bilevel_grad}
\end{align}
Because the derivative of the sign function, $\texttt{sgn}'(\BZ)$, is zero almost everywhere, the exact gradient vanishes, making gradient-based optimisation impossible.\footnote{Similar to many common activation functions in Deep Learning, like ReLU, the sign function is not differentiable at $x=0$. We define $\texttt{sgn}'(0)=0$.} 
The STE addresses this by replacing the true derivative $\texttt{sgn}'(\BZ)$ with a non-zero surrogate gradient $g^\text{STE}(\BZ)$ during the backward pass
\begin{align}
    \nabla_\BZ^\text{STE} \mathcal{F}(\BZ, \alpha) = g^\text{STE}(\BZ) \odot \nabla_\BA \mathcal{F}(\BA, \alpha).
\end{align}
Surrogate gradient is applied element-wise to the entries of $\BZ$.
Common choices include
\begin{itemize}
    \item Identity STE: $g^\text{id}(z) = 1$,
    \item Clipped STE: $g^\text{clipped}(z) = \frac{1}{a} \mathbf{1}_{|z| \le a}$, for $a>0$,
    \item Tanh STE: $g^\text{tanh}(z) = \beta (1 - \tanh^2(\beta z))$, for $\beta>0$.
\end{itemize}

\section{Probabilistic Interpretation of the STE}
\label{sec:prob_inter_ste}
Although the STE is often presented as a heuristic to bypass zero gradients, it can also be interpreted as the exact gradient of a smoothed surrogate objective. 
Specifically, for a scalar $z \in \mathbb{R}$ and $h(z) = \texttt{sgn}(z)$, consider the smoothed function
\begin{align}
    \tilde{h}(z) = \mathbb{E}_{\epsilon \sim p(\epsilon)}[h(z + \epsilon)].
\end{align}
That is, $\tilde h$ is the expectation of $h$ under additive noise $\epsilon \sim p(\epsilon)$
This corresponds to convolving the $\texttt{sgn}$ function with the density $p(\epsilon)$, yielding a differentiable \text{soft} approximation. 
As shown below, under this viewpoint, common STE choices arise naturally from specific noise distributions $p(\epsilon)$. 
\begin{lemma}[Uniform Noise]
    Let $h(z) = \texttt{sgn}(z)$ and $\epsilon \sim \mathcal{U}[-a, a]$ for $a > 0$. Then,  $\tilde{h}'(z)=g^\text{clipped}(z)$.
\end{lemma}

\begin{proof}
We obtain 
\begin{align}
    \tilde{h}(z) =\frac{1}{2a} \int_{-a}^{a} \texttt{sgn}(z + \epsilon) d\epsilon = \begin{cases}
        1, \quad \text{if } z > a \\ 
        -1, \quad \text{if } z < -a \\ 
        z/a, \quad \text{if } |z| \le a \\ 
    \end{cases},
\end{align}
Computing the derivative completes the proof
\begin{align}
    \tilde{h}'(z) = 
    \begin{cases} 
        \frac{1}{a} & \text{if } |z| \le a \\ 
        0 & \text{if } |z| > a 
    \end{cases} 
    = g^\text{clipped}(z).
\end{align}
\end{proof}

\begin{lemma}[Logistic Noise]
Let $h(z) = \texttt{sgn}(z)$ and  $\epsilon \sim \text{Logistic}(0, s)$ with scale $s > 0$. Then, $\tilde{h}'(z)=g^\text{tanh}(z)$.
\end{lemma}

\begin{figure}[t]
    \centering
    \includegraphics[width=1.0\linewidth]{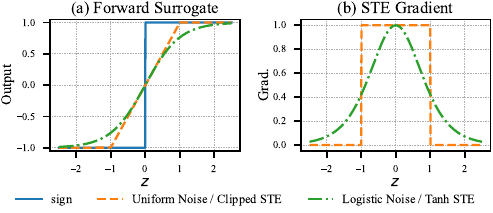}
    \caption{The Tanh and Clipped STE with the uniform and logistic surrogate.}
    \label{fig:ste_analysis}
\end{figure}

\begin{proof}
    Recall that $\texttt{sgn}(x) = 2H(x) - 1$, where $H(x)$ is the Heaviside step function. 
    We thus get
    \begin{align}
        \tilde{h}(z) = \E_\epsilon [2H(z+\epsilon) - 1] = 2 P(z + \epsilon \ge 0) - 1.
    \end{align}
    The symmetry of the logistic distribution now yields
    \begin{align}
        P(z + \epsilon \ge 0) = P(\epsilon \ge -z) = 1 - F_{\text{logistic}}(-z),
    \end{align}
    where $F_{\text{logistic}}(x) = \frac{1}{1 + e^{-x/s}}$. Thus
    \begin{align}
        P(z + \epsilon \ge 0) = 1 - \frac{1}{1 + e^{z/s}} = \frac{e^{z/s}}{1 + e^{z/s}} = \sigma\left(\frac{z}{s}\right),
    \end{align}
    where $\sigma(\cdot)$ is the sigmoid function. Substituting back into $\tilde{h}(z)$
    \begin{align}
        \tilde{h}(z) = 2 \sigma\left(\frac{z}{s}\right) - 1 = \tanh\left(\frac{z}{2s}\right).
    \end{align}
    Differentiating with respect to $z$
    \begin{align}
        \tilde{h}'(z) &= \frac{d}{dz} \tanh\left(\frac{z}{2s}\right) = \frac{1}{2s} \left( 1 - \tanh^2\left(\frac{z}{2s}\right) \right).
    \end{align}
    Setting $\beta = \frac{1}{2s}$, we recover the form of the Tanh STE, i.e., $g^\text{tanh}(z) = \beta(1 - \tanh^2(\beta z))$.
\end{proof}
In Fig.~\ref{fig:ste_analysis}, we illustrate these two surrogates and the resulting STE gradients.
The parameters $a$ (for uniform noise) and $\beta$ (for logistic noise) control the extent of gradient. 
Increasing the noise scale spreads the gradient over a wider range of the latent space, helping to prevent the optimiser from getting stuck in regions where the binary output is locally constant. 
The Tanh STE has infinite support (right panel of Fig.~\ref{fig:ste_analysis}), ensuring that even latent values far from the decision boundary receive a small, non-zero gradient.

Interestingly, the identity STE $g^\text{id}$ cannot be obtained from this smoothing perspective.
For $\tilde{h}'\equiv g^{\text id}\equiv 1$ to hold would require a noise density $p(\epsilon)$ that is constant over the entire real line, which is impossible for a valid probability distribution. 

\section{Numerical Experiments}
\label{sec:numerical_exp}
We evaluate the proposed bilevel framework for pattern optimisation in terms of reconstruction quality across sampling ratios and robustness to noise.
We also compare against end-to-end deep learning frameworks, and empirically analyse the STE solution with respect to the global optimum. All experiments are conducted on cell microscopy images from CytoImageNet~\cite{Hua_2021_CytoImageNetLargescalePretraining}. 

\subsection{Experimental Setup}
\paragraph{Dataset}
We validate our approach on a subset of the CytoImageNet dataset \cite{Hua_2021_CytoImageNetLargescalePretraining}, a large-scale collection of cell microscopy images consisting of $890$K images. To mimic the limited-data regime common in biomedical imaging, we use only \num{1000} images to optimise the illumination patterns, and use an independent set of $100$ images for evaluation. For comparisons with end-to-end deep learning methods, we consider training subsets ranging from \num{250} to \num{25000} images. All images are resized to $128\times128$ px$^2$, consistent with physical constraints in single-pixel microscopy~\cite {Ebner_2023_DiffractionlimitedHyperspectralMidinfrared}. Images with a BRISQUE score~\cite{Mittal_2012_NoReferenceImageQuality} above $25.0$ are discarded to exclude uninformative or corrupted samples.

\paragraph{Baseline Methods} We consider two fixed pattern baselines: random Gaussian, the standard choice in compressive sensing, and Scrambled Hadamard (SH). In both cases, image reconstruction is performed with Total Variation (TV) regularisation, solved with the Primal-Dual Hybrid Gradient algorithm (PDHG)~\cite{Chambolle_2011_FirstOrderPrimalDualAlgorithm}, yielding \emph{Gaussian--TV} and \emph{SH--TV} baselines. For comparisons against end-to-end methods, we consider DCAN~\cite{Higham_2018_DeepLearningRealtime}, the standard deep learning method for pattern optimisation in SPI. 

\paragraph{Learned Regularisers} We replace TV with two learned regularisers: TDV~\cite{Kobler_2020_TotalDeepVariation} ($\approx 400$K parameters) and WCRR~\cite{Goujon_2023_LearningWeaklyConvex} ($\approx 15$K parameters).
Both are pretrained on the BSDS500 dataset~\cite{Martin_2001_DatabaseHumanSegmented} for denoising, independently of the forward operator, following the protocol in \cite {Hertrich_2025_LearningRegularizationFunctionals}. 
WCRR is significantly lighter than TDV in terms of number of parameters, offering a trade-off between computational cost and reconstruction quality.

\paragraph{Lower-Level Solver}
To solve the lower-level problem~\eqref{eq:bilevel_problem2} we employ the nonmonotonic Accelerated Proximal Gradient algorithm (nmAPG)~\cite{Li_2015_AcceleratedProximalGradient}. 
This choice is motivated by the potential non-convexity of the chosen regulariser $\mathcal{J}$, which would preclude standard choices such as proximal gradient descent and accelerated variants with a monotonic line search, as they would fail to converge to a critical point. 

\paragraph{Implementation Details}
Experiments are run on an NVIDIA GeForce RTX 4070 Ti SUPER GPU. 
Unless stated otherwise, as the noise process $P$ we assume additive Gaussian noise with standard deviation $\sigma=5.0$, during both training and testing. 
Patterns are trained for $10$ epochs, with the lower level solver (nmAPG) limited to $1000$ iterations. 
We use Tanh-STE with slope $\beta = 1.0$, unless stated otherwise, and a batch size of $50$ images for the bilevel methods.
For baseline methods, PDHG is run for $150$ iterations, with $300$ iterations for the computation of the TV proximal operator.
The code is available at \href{https://github.com/CrisSherban/sppo}{https://github.com/CrisSherban/sppo}.

\subsection{Reconstruction Quality of Learned Patterns}

\begin{table*}[t]
\centering
\caption{Reconstruction quality (PSNR/SSIM) for different numbers of patterns $M$. Learned approaches are trained on \num{1000} images.}
\label{tab:psnr_ssim_grouped}
\setlength{\tabcolsep}{3pt}
\begin{tabular}{l cc c cc c cc c cc c cc}
\toprule
 & \multicolumn{2}{c}{$M=128$ (0.8\%)} & \hspace{8pt}
 & \multicolumn{2}{c}{$M=256$ (1.6\%)} & \hspace{8pt}
 & \multicolumn{2}{c}{$M=512$ (3.1\%)} & \hspace{8pt}
 & \multicolumn{2}{c}{$M=1024$ (6.3\%)} & \hspace{8pt}
 & \multicolumn{2}{c}{$M=2048$ (12.5\%)} \\
\cmidrule(lr){2-3} \cmidrule(lr){5-6} \cmidrule(lr){8-9} \cmidrule(lr){11-12} \cmidrule(lr){14-15} 
& PSNR ($\uparrow$) & SSIM ($\uparrow$) &
& PSNR ($\uparrow$) & SSIM ($\uparrow$) &
& PSNR ($\uparrow$) & SSIM ($\uparrow$) &
& PSNR ($\uparrow$) & SSIM ($\uparrow$) &
& PSNR ($\uparrow$) & SSIM ($\uparrow$)  \\
\midrule
Gaussian--TV    & 19.95 & 0.464 && 22.14 & 0.528 && 24.81 & 0.606 && 27.52 & 0.691 && 30.30 & 0.778 \\
SH--TV          & 19.99 & 0.457 && 22.29 & 0.523 && 24.85 & 0.601 && 27.60 & 0.689 && 30.43 & 0.782 \\
\midrule
Gaussian--WCRR  & 19.17 & 0.461 && 22.59 & 0.547 && 25.77 & 0.640 && 27.64 & 0.698 && 29.37 & 0.754 \\
Gaussian--TDV   & 21.09 & 0.501 && 23.94 & 0.581 && 26.02 & 0.643 && 27.90 & 0.704 && 29.61 & 0.760 \\
SH--WCRR        & 19.19 & 0.461 && 23.04 & 0.565 && 25.77 & 0.640 && 27.71 & 0.700 && 29.44 & 0.756 \\
SH--TDV         & 21.61 & 0.514 && 24.02 & 0.585 && 26.07 & 0.647 && 27.97 & 0.707 && 29.72 & 0.764 \\
\midrule
DCAN            & 22.40 & 0.513 && 22.42 & 0.509 && 22.73 & 0.516 && 22.66 & 0.516 && 22.73 & 0.511 \\
\midrule
RnP--WCRR       & 24.97 & 0.611 && 26.82 & 0.663 && 28.17 & 0.706 && 29.42 & 0.748 && 30.80 & 0.793 \\
RnP--TDV        & 25.30 & 0.618 && 26.88 & 0.666 && 28.00 & 0.701 && 29.20 & 0.743 && 30.49 & 0.783 \\
STE--WCRR       & 24.98 & 0.611 && 27.36 & 0.678 && 29.02 & 0.732 && 30.41 & 0.778 && \bf{31.42} & \bf{0.812} \\
STE--TDV        & \bf{25.43} & \bf{0.625} && \bf{27.44} & \bf{0.682} && \bf{29.10} & \bf{0.736} && \bf{30.51} & \bf{0.783} && 31.22 & 0.806 \\
\bottomrule
\end{tabular}
\end{table*}

\begin{figure*}[t]
    \centering
    \includegraphics[width=0.95\linewidth]{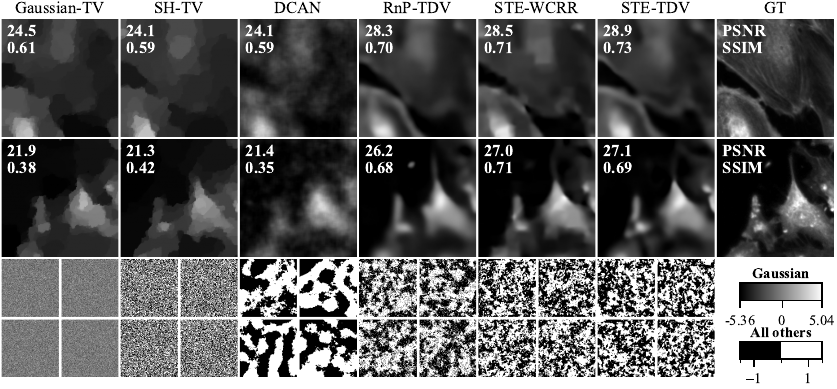}
    \caption{Reconstruction of test images comparing baselines and learned approaches at $M=256$ trained with \num{1000} images. The last row shows the first $4$ patterns from each method.}
    \label{fig:performance_overall_images}
\end{figure*}

\begin{figure}[t]
    \centering
    \includegraphics[width=0.95\linewidth]{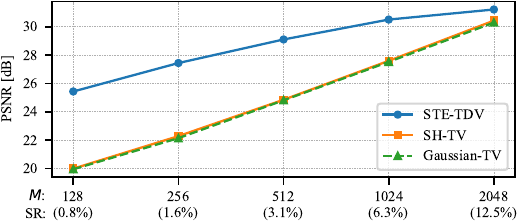}
    \caption{Reconstruction PSNR at various sampling ratios comparing STE-TDV to the considered baselines.}
    \label{fig:performance_learned_psnr}
\end{figure}

We present three experiments. The first focuses on the benefits of replacing TV with a learned regulariser under classical patterns. 
The second explores the additional quality increase from optimising the pattern via the bilevel formulation~\eqref{eq:bilevel_problem1}. 
In the last experiment, we investigate the benefits of using STE as the binarisation mechanism. 
For learned methods, patterns are optimised independently for each sampling ratio~$M$, following the DCAN protocol. 
Quantitative results are provided in Table~\ref{tab:psnr_ssim_grouped}. 
Qualitative results are shown in Fig.~\ref{fig:performance_overall_images}, where we present reconstruction (and patterns) at $M=256$.

\paragraph{Effect of learned regularisers with classical patterns} Replacing TV with WCRR or TDV under fixed Gaussian or SH pattern yields consistent improvements in PSNR and SSIM across all sampling ratios. Among fixed designs, SH patterns marginally outperform the Gaussian ones. TDV provided a larger performance increase in highly undersampled regimes, whereas the performance of the two learned regularisers converges in performance as $M$ increases.

\paragraph{Effect of learned patterns} Optimising the illumination patterns via the bilevel formulation leads to substantial further improvements, with gains of up to $4$dB in PSNR over the fixed-pattern baselines, see also Fig.~\ref{fig:performance_learned_psnr}. This improvement is most pronounced at low sampling ratios: STE--TDV at $M=256$ achieved a PSNR comparable to SH--TV at $M=1024$, suggesting that the learned patterns can reduce the required number of measurements by a factor of four without loss of reconstruction quality. Qualitatively, the WCRR tends to reconstruct slightly blockier images (see Fig.~\ref{fig:performance_overall_images}), with the gain of speed and lighter memory requirements. The optimised patterns exhibit similar structure after training with both WCRR and TDV, see the last row in Fig.~\ref{fig:performance_overall_images}.

\begin{figure}[t]
    \centering
    \includegraphics[width=0.95\linewidth]{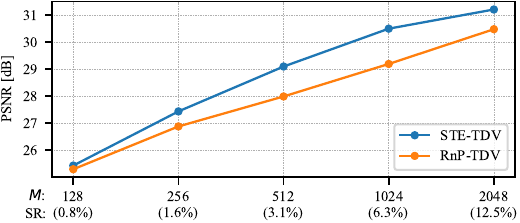}
    \caption{Reconstruction PSNR at various sampling ratios comparing the choice of the binarisation strategy: RnP vs STE.}
    \label{fig:bin_strategy}
\end{figure}

\paragraph{Effect of binarisation strategies} Both binarisation strategies, STE and RnP \eqref{eq:rnp}, outperform the fixed-pattern methods. Here, STE achieves slightly higher PSNR and SSIM than RnP, particularly at low sampling ratios $M$. RnP exhibits more unstable behaviour on test scenarios because the reconstructor receives measurements from an exact binary sampling matrix at test time, while the underlying sampling matrix is only approximately binary at training time. Additionally, it is difficult to achieve a perfect scheduler for the binary penalty~\cite{Lucidi_2010_ExactPenaltyFunctions}. Although the relaxed RnP problem has the same minimiser as its corresponding binary problem, we empirically notice that the STE surrogates provide better reconstructions for all sampling ratios, see Fig.~\ref{fig:bin_strategy}.

\begin{figure*}[t]
    \centering
    \includegraphics[width=0.95\linewidth]{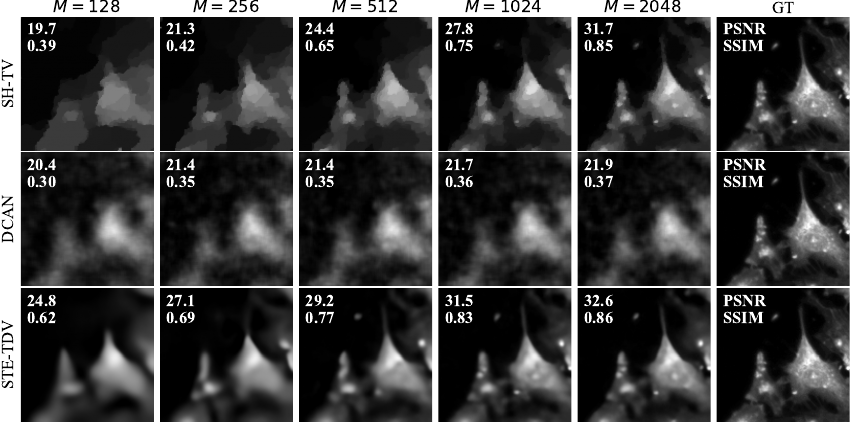}
    \caption{Reconstruction of test images across different sampling ratios for SH-TV, DCAN, and STE-TDV. The learned methods are trained with $1000$ images.}
    \label{fig:performance_across_m_images}
\end{figure*}

Reconstructions across sampling ratios are shown in 
Fig.~\ref{fig:performance_across_m_images}. 
We observe that STE--TDV provides qualitatively and quantitatively better reconstructions, particularly in extreme undersampling regimes ($M = 128$, $M = 256$). Where baseline methods create aberrations or artefacts, STE--TDV provides well-maintained cellular structures.

\subsection{Generalisation to Unseen Regularisers}
\begin{figure}[t]
    \centering
    \includegraphics[width=0.95\linewidth]{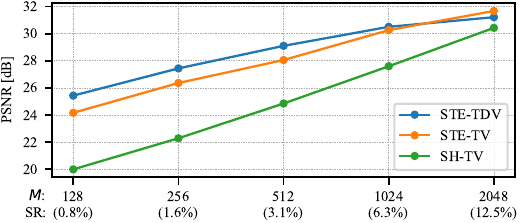}
    \caption{Reconstruction PSNR comparing the effects of unseen regularisers.}
    \label{fig:regularisation_generalisation}
\end{figure}

We examine the sensitivity of the learned patterns to the choice of the regulariser by testing the patterns optimised using STE-TDV with TV reconstruction at test time (cf. Fig.~\ref{fig:regularisation_generalisation}). 
We observe that this mismatch leads to a noticeable degradation of reconstruction quality in highly undersampled regimes, although the performance still exceeds the baselines. 
As the sampling ratio increases beyond $\approx 6.3\%$, the gap becomes negligible, highlighting stronger coupling between the optimised patterns and the regulariser at low sampling ratios.

\subsection{Robustness to Misspecified Noise Level}

\begin{figure}[t]
    \centering
    \includegraphics[width=0.99\linewidth]{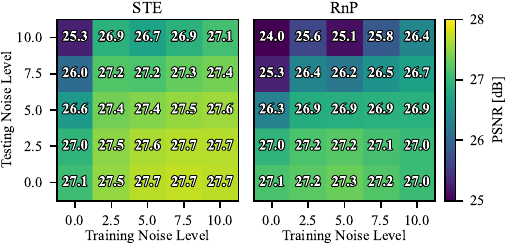}
    \caption{Reconstruction PSNR for training noise with respect to testing noise standard deviations, for STE-TDV (left) and RnP-TDV (right) at $M=256$.}
    \label{fig:noise_analysis_ste_rnp}
\end{figure}
We investigate the robustness of the learned patterns to noise-level mismatches between training and test time, by considering $P_{\sigma}(\By) = \By + \boldsymbol{\xi},$ where $\boldsymbol{\xi} \sim \mathcal{N}(\mathbf{0}, \sigma^2 \mathbf{I}_N)$ and $\sigma\in\{0,2.5, 5, 7.5. 10\}$.
As expected, increasing noise at test time degrades the reconstruction quality (column-wise trends in Fig.~\ref{fig:noise_analysis_ste_rnp}).
However, rather than peaking when the training and test noise levels coincide, STE-trained patterns perform best when the training noise level is higher than the test time noise level.
This highlights that robustness rather than noise-specific tuning is the primary driver of performance.
In data-driven inverse problems, such behaviour is often expected~\cite{Krainovic_2023_LearningProvablyRobust}. 

We observe similar trends for RnP learned patterns. 
However, for some noise levels, training with larger noise levels ($\sigma\geq5.0$) leads to a degradation in performance, making STE the more robust choice. 

\subsection{Comparison with end-to-end deep learning}
\begin{figure*}[t]
    \centering
    \includegraphics[width=0.95\linewidth]{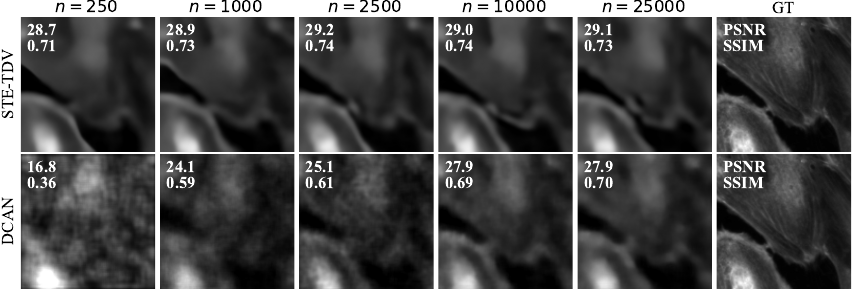}
    \caption{Reconstructed images at different training dataset sizes ($n$) for STE-TDV and DCAN ($M=256$).}
    \label{fig:dcan_vs_bilevel_images}
\end{figure*}

\begin{figure}[t]
    \centering
    \includegraphics[width=0.95\linewidth]{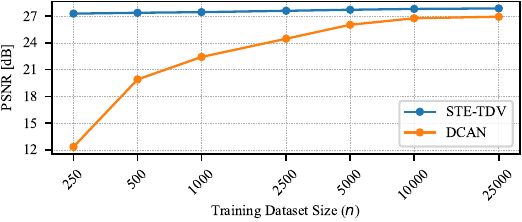}
    \caption{Reconstruction PSNR at various training set sizes, comparing STE-TDV to DCAN for $M=256$.}
    \label{fig:dcan_vs_bilevel}
\end{figure}

End-to-end deep learning methods such as DCAN~\cite{Higham_2018_DeepLearningRealtime} jointly learn illumination patterns and the reconstruction network, achieving strong performance when sufficient training data is available, but degrading sharply in data-scarce regimes. 
To investigate this effect, we re-train both STE--TDV and DCAN on CytoImageNet with training set sizes ranging from \num{250} to \num{25000} images. 
Example reconstructions at different training set sizes are shown in Fig.~\ref{fig:dcan_vs_bilevel_images} with PSNR reported in Fig.~\ref{fig:dcan_vs_bilevel}. 

Our approach outperforms DCAN across all training set sizes and is particularly advantageous in low-data regimes. 
In particular, under the default training set setting (\num{1000} images), we observe that the performance of DCAN is largely unaffected by the sampling ratio $M$, cf. Table~\ref{tab:psnr_ssim_grouped}.
The learned patterns also differ substantially. 
In particular, patterns learned with our method capture higher spatial frequencies (Fig.~\ref{fig:performance_overall_images}), whereas DCAN patterns concentrate on lower frequencies. 

\subsection{Choice of STE Surrogate}
In line with prior theoretical results on gradient estimators~\cite{Schoenbauer_2024_CustomGradientEstimators}, we find that the specific choice of STE has a negligible impact on reconstruction quality. 
We provide results for different choices of the slope ($\beta$) for Tanh STE and the threshold ($a$) for Clipped STE in Table~\ref{tab:ste_comparison}. 
Given the lack of observed differences in training stability, we select the Tanh STE with $\beta = 1.0$ as the default configuration.

\begin{table}[t]
\centering
\caption{PSNR comparison for Tanh STE and Clipped STE with different parameters, with $M=512$. The identity STE provides a reconstruction PSNR of $29.13$ dB.}
\label{tab:ste_comparison}
\begin{tabular}{lcccccc}
\toprule
Parameter & 0.25 & 0.50 & 0.75 & 1.00 & 1.25 & 1.50 \\
\midrule
Tanh STE ($\beta$)   & 28.98 & 29.10 & 29.11 & 29.10 & 29.07 & 29.08 \\
Clipped STE ($a$)    & 28.98 & 29.08 & 29.08 & 29.13 & 29.11 & 29.11 \\
\bottomrule
\end{tabular}
\end{table}

\subsection{STE Solutions vs.\ Global Optima}
Since exhaustive search over all binary sensing matrices is tractable only at very small scales, we compare STE solutions to global optima on $3\times3$ images with $M=3$ measurements. 
We simplify the bilevel problem to a noiseless single-image setting
\begin{equation}
  \min_{\mathbf{Z} \in \mathbb{R}^{M \times N}} \;
  \|\mathbf{x} - \hat{\mathbf{x}}\!\left(\texttt{sgn}(\mathbf{Z});\,
  \mathbf{y}\right)\|_2^2,
   \ \ \text{s.t.} \ \ 
  \hat{\mathbf{x}}(\mathbf{A};\mathbf{y}) = \mathbf{A}^{\dagger}\mathbf{y},
  \label{eq:bilevel_reduced}
\end{equation}
where $\mathbf{A}^{\dagger}$ denotes the pseudoinverse. 
For each of $30$ random images, we compute the STE solution and compare it against the ground truth by exhaustively evaluating all $2^{MN}$ binary sensing matrices to identify the global minimiser. 
When multiple global optima exist, we select the one closest to the STE solution in cosine distance.

\begin{figure}[t]
    \centering
    \includegraphics[width=0.95\linewidth]{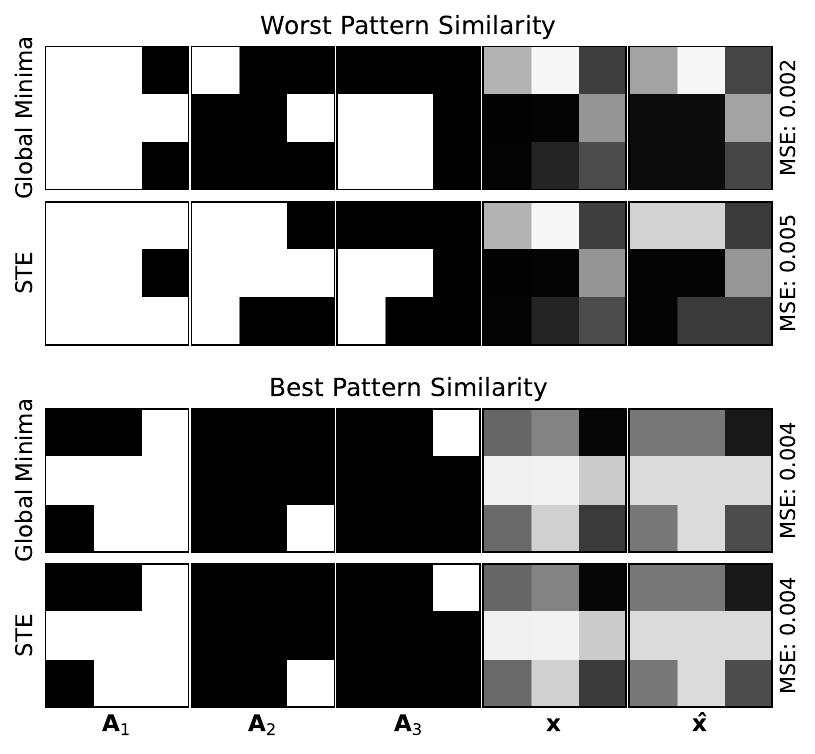}
    \caption{STE solution compared to the closest global minima for $3 \times 3$ images. Columns $1-3$ show the optimised patterns, column $4$ the image, and column $5$ the minimum norm solution.}
    \label{fig:bruteforce_comparisons}
\end{figure}

Fig.~\ref{fig:bruteforce_comparisons} shows the worst and best solutions according to the cosine distance between the closest global minima and the STE solution. 
In most cases, the STE solution aligns closely with a global optimum. 
In failure cases, it converges to a local minimum, but the resulting reconstructions incur only a marginal increase in MSE. 
This confirms that STE solutions are reliable surrogates, either recovering global optima or producing near-optimal solutions at local minima.

\section{Discussion and Conclusions}\label{sec:conc}
This paper presents a bilevel framework designed for learning binary illumination patterns in SPI under noisy conditions. 
Rather than relying on a traditional TV regulariser, we propose using learned regularisers such as WCRR and TDV. 
Our findings demonstrate that utilising these learned regularisers enhances reconstruction performance for classical patterns. 
When we additionally learn the sampling patterns, we observe even greater improvements in reconstruction quality.
The inherent characteristics of binary patterns pose significant optimisation challenges. 
We show that the STE plays a vital role in overcoming these challenges, providing a robust solution for managing constraints in SPI. 
Furthermore, our approach leads to substantial performance gains, even with small training set sizes, achieving impressive results with as few as 250 images, all while maintaining high reconstruction quality.

A promising area for future work is to adapt pre-trained regularisers during the optimisation of patterns. 
It would also be valuable to evaluate learned patterns on real SPI systems and downstream tasks (e.g., data fusion).
Additionally, exploring adaptive or sequential pattern learning could further improve reconstruction quality in highly undersampled regimes.

\bibliographystyle{IEEEtran}
\bibliography{bibliography}

\end{document}